%% file: main.tex
\documentclass[10pt,twocolumn,letterpaper]{article}

\usepackage{cvpr}              % To produce the CAMERA-READY version
\usepackage{booktabs}   % 优雅的表格线
\usepackage{multirow}   % 允许单元格跨多行
\usepackage[table]{xcolor} % 行/列着色
\usepackage{pifont}        % 勾选符号
\usepackage{graphicx}
\usepackage{caption}
\usepackage{subcaption}
\usepackage[accsupp]{axessibility}  % Improves PDF readability for those with disabilities.
\newcommand{\cmark}{\ding{51}}
\newcommand{\xmark}{--}
\input{preamble}
\definecolor{cvprblue}{rgb}{0.21,0.49,0.74}
\usepackage[pagebackref,breaklinks,colorlinks,allcolors=cvprblue]{hyperref}

\def\paperID{35491} % *** Enter the Paper ID here
\def\confName{CVPR}
\def\confYear{2026}

\title{Dual-level Adaptation for Multi-Object Tracking: 

Building Test-Time Calibration from Experience and Intuition}

\author{
Wen Guo$^1$, Pengfei Zhao$^1$, Zongmeng Wang$^4$, 
Yufan Hu$^{2}$\thanks{Corresponding Author.}, Junyu Gao$^3$\\
$^1$Shandong Technology and Business University, $^2$University of Science and Technology Beijing, \\ 
$^3$Institute of Automation, Chinese Academy of Sciences, $^4$Inner Mongolia University,\\
{\tt\small \{wguo,2024420014\}@sdtbu.edu.cn, wangzongmeng612@gmail.com}\\
{\tt\small huyufanqaixuan@gmail.com, junyu.gao@nlpr.ia.ac.cn}
}
\begin{document}
\maketitle

\input{sec/0_abstract}    
\input{sec/1.Introduction.tex}
\input{sec/2.Related_Work.tex}
\input{sec/3.Method.tex}
\input{sec/4.Experiments.tex}
\input{sec/5.Conclusion.tex}
\section*{Acknowledgments}
\input{sec/Acknowledgements.tex}
{
    \small
    \bibliographystyle{ieeenat_fullname}
    \bibliography{main}
}

% WARNING: do not forget to delete the supplementary pages from your submission 
% \input{sec/X_suppl}

\end{document}

%% file: sec/0_abstract.tex
\begin{abstract}
Multiple Object Tracking (MOT) has long been a fundamental task in computer vision, with broad applications in various real-world scenarios.
However, due to distribution shifts in appearance, motion pattern, and catagory between the training and testing data, model performance degrades considerably during online inference in MOT.
Test-Time Adaptation (TTA) has emerged as a promising paradigm to alleviate such distribution shifts.
However, existing TTA methods often fail to deliver satisfactory results in MOT, as they primarily focus solely on frame-level adaptation while neglecting temporal consistency and identity association across frames and videos.
Inspired by human decision-making process, this paper propose a Test-time Calibration from Experience and Intuition (TCEI) framework.
In this framework, the Intuitive system utilizes transient memory to recall recently observed objects for rapid predictions, while the Experiential system leverages the accumulated experience from prior test videos to reassess and calibrate these intuitive predictions.
Furthermore, both confident and uncertain objects during online testing are exploited as historical priors and reflective cases, respectively, enabling the model to adapt to the testing environment and alleviate performance degradation.
Extensive experiments demonstrate that the proposed TCEI framework consistently achieves superior performance across multiple benchmark datasets and significantly enhances the model's adaptability under distribution shifts.
The code will be released at \url{https://github.com/1941Zpf/TCEI}.
\end{abstract}

%% file: sec/1.Introduction.tex
\section{Introduction}
\label{sec:Introduction}

\begin{figure}
  \centering
   \includegraphics[width=0.95\linewidth]{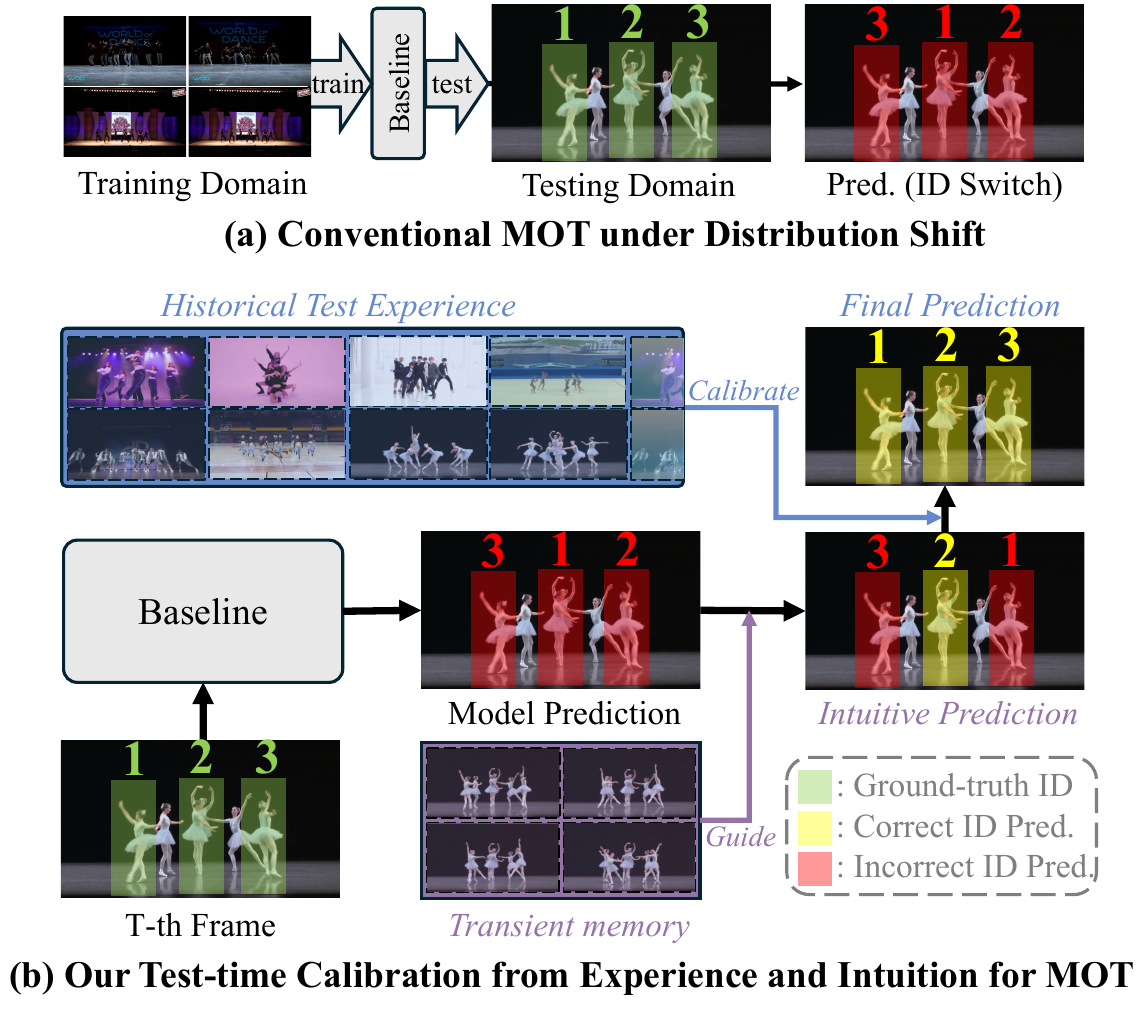}
   \caption{
            % Illustration of the proposed Test-time Calibration from Experience and Intuition (TCEI) framework.
            % The bottom part (\textit{w/o TTA}) shows the baseline model directly performing inference without test-time adaptation, resulting in incorrect predictions under distribution shift.
            % The upper part depicts our TCEI framework, where the \textit{Intuitive System} leverages transient memory to guide predictions, and the \textit{Experiential System} further calibrates them using historical test experience, ultimately producing more accurate and robust final predictions.
            Illustration of the proposed TCEI framework.
            (a) The upper part illustrates MOT under distribution shift. 
            Due to significant discrepancies between the training and testing domains, the baseline model produces incorrect ID predictions.
            (b) The lower part presents our TCEI framework. The Intuitive System exploits transient memory from recently observed objects to provide rapid test-time guidance, and the Experiential System utilizes accumulated historical experience to calibrate these intuitive predictions. 
            % By combining short-term cues with long-range experience, TCEI produces more accurate and robust identity predictions under distribution shift.
            }
  \label{fig:illustration}
\end{figure}

Multiple Object Tracking (MOT) aims to detect objects of interest in continuous video sequences and associate identical objects across frames into coherent trajectories~\cite{bergmann2019tracking,wojke2017simple2,gao2017deep_relative_tracking,gao2019graph}. 
As a fundamental task in computer vision, it has been widely applied in various real-world scenarios such as intelligent surveillance, autonomous driving, and sports analytics~\cite{hu2019joint,cui2023sportsmot,ristani2018features}. 
However, the inherent complexity and stochasticity of real-world environments often lead to distribution shifts between the training and inference data, including category~\cite{li2023ovtrack}, appearance shifts~\cite{segu2023darth}, and motion pattern shifts~\cite{sun2022dancetrack}, which cause trained models to encounter unseen or out-of-distribution scenarios during testing. 

Test-Time Adaptation (TTA) mitigates performance degradation caused by data shifts by dynamically adjusting model parameters and prediction outputs based on unlabeled test data. 
However, existing TTA methods are primarily applied to static image tasks such as image classification~\cite{wang2020tent} and semantic segmentation~\cite{shin2022mm}. 
Although some methods~\cite{gao2023fast,shao2025pura} have attempted to extend TTA to dynamic image processing, they still lack the ability for temporal modeling in complex scenes, as they typically rely only on intra-frame information for processing.
However, in multi-object tracking, both intra-frame and inter-frame information are important because intra-frame cues help distinguish objects within a single frame, whereas inter-frame temporal cues ensure identity(ID) consistency across time.

Daniel Kahneman's dual-system theory~\cite{kahneman2011thinking} provides valuable insights into how humans achieve temporal modeling during the decision-making process. 
Human decision-making initially relies on intuitive, automatic processes that draw upon associative and transient memory to generate rapid judgments about the current problem.
These intuitive judgments are subsequently monitored and adjusted by reflective, experience-based reasoning processes that provide deliberate evaluation and correction.
Inspired by this theory, as shown in \cref{fig:illustration}, we propose a Test-time Calibration from Experience and Intuition (TCEI) framework. 

Specifically, the Intuitive system first considers guiding the model by recalling recently processed objects, and it therefore constructs a transient memory to store these recent objects.
The recent objects with confident predictions are employed as temporal priors to enhance the accuracy of the current predictions.
To further enhance prediction robustness, we introduce a reflection mechanism guided by the recent objects with uncertain predictions. 
These recent uncertain objects serve as reflective cases to prompt the model to reassess and avoid making similarly unreliable predictions. 
The Intuitive system enables the model to generate more comprehensive and adaptive predictions by combining the knowledge learned during training with the transient memory derived from recently processed objects.
However, because the Intuitive system only recalls recently processed objects, it cannot provide long-range temporal information. 
To address this limitation, we construct the Experiential system, which leverages the knowledge accumulated from all previously processed videos to validate and calibrate the intuitive predictions.
When the intuitive predictions are consistent with historical experience, the Experiential system refrains remains inactive to preserve the stability of intuitive inference.
However, when discrepancies arise between the intuitive predictions and prior experience, the Experiential system actively engages to calibrate the intuitive outputs.
% During inference, TCEI generates intuitive predictions through the Intuitive system and then refines them using the Experiential system, yielding more reliable final outputs.
It is worth noting that TCEI is a forward-propagation-based TTA method that requires no additional training or backpropagation. 
Consequently, this human-inspired framework enables timely predictions and improves the model's robustness under distribution shifts.
We validate the effectiveness of our approach on multiple benchmark datasets, where it consistently achieves superior performance. 
Our main contributions are summarized as follows:
\begin{itemize}
    \item We propose a test-time calibration from experience and intuition framework for MOT. It leverages historical objects observed in the testing environment to reassess and calibrate the tracker's current ID predictions, thereby enhancing the robustness of MOT under online test-time distribution shifts.
    \item We further exploit both transient memory from recently observed objects and experience accumulated from previously processed test videos to provide adaptive guidance for ID association in MOT, effectively mitigating performance degradation caused by appearance variations, motion irregularities, and other distribution shifts in test data.
    \item Experimental results on three mainstream datasets demonstrate that our method exhibits superior performance and strong generalization capability.
\end{itemize}

%% file: sec/2.Related_Work.tex
\section{Related Work}
\label{sec:Related_Work}

{\bf Multi-Object Tracking.}
MOT aims to continuously identify and associate the trajectories of multiple objects within video sequences, representing a core and fundamental problem in the field of computer vision. 
In recent years, research has primarily focused on tracking-by-detection paradigms and end-to-end architectures based on Transformers.

Tracking-by-Detection paradigm~\cite{bewley2016simple,wojke2017simple} has long served as the fundamental paradigm for MOT. 
% This paradigm divides the MOT task into two stages: in the detection stage, all objects of interest are detected within each single frame; in the association stage, objects with the same identity across different frames are linked into trajectory segments. 
The strong baseline model ByteTrack~\cite{zhang2022bytetrack} adopts YOLOX~\cite{ge2021yolox} as its detector in the detection stage and employs Kalman filtering~\cite{welch1995introduction} for motion estimation during the association stage. 
% It performs a two-stage association by using intersection-over-union (IoU) and the Hungarian algorithm~\cite{kuhn1955hungarian} to successively associate high-confidence and low-confidence detections. 
Subsequent works have proposed various improvements. 
For example, OC-SORT~\cite{cao2023observation} introduces an observation-centric association strategy; 
Deep OC-SORT~\cite{maggiolino2023deep} incorporates ReID-based appearance features; 
and Hybrid-SORT~\cite{yang2024hybrid} adds Tracklet Confidence Modeling (TCM) and Height-Modulated IoU (HM-IoU) for enhanced robustness. 
However, tracking-by-detection paradigms suffer from inherent limitations, such as the inability to model long-range dependencies and the reliance on heuristic matching, making it difficult to learn complex cross-frame relationships.

In recent years, with the emergence of the DETR series of detection models~\cite{carion2020end,zhu2020deformable}, Transformer-based approaches have become another mainstream paradigm, leveraging their end-to-end advantage and the Transformer's powerful sequence modeling and global attention mechanisms~\cite{vaswani2017attention}. 
MOTR~\cite{zeng2022motr} introduces the concept of track queries for trajectory association, achieving a fully end-to-end MOT framework. 
MOTRv2\cite{zhang2023motrv2} further improves performance by incorporating an external detector. 
The latest method, MOTIP~\cite{gao2025multiple}, reformulates the association process as a direct ID prediction task through an ID decoder. 
% By propagating ID information from historical trajectories directly to new detections, MOTIP effectively addresses the conflict issue in MOTR, which arises from the over-coupling between detection and association tasks.

Although current MOT methods have achieved remarkable results, they still face the distribution shift problem~\cite{khosla2012undoing}, where the distribution of training data differs from that of testing data. 
This discrepancy leads to performance degradation during testing or real-world deployment.

\noindent {\bf Test-Time Adaptation.}
TTA~\cite{liang2025comprehensive} aims to adaptively optimize model predictions during the testing phase using only test samples, enabling the model to adjust to test data that may differ in distribution from the training data. 
In recent years, TTA has been widely applied in the vision-language model domain~\cite{radford2021learning,gao2025learning,gao2020learning,gao2023vectorized}, establishing a unified framework for lightweight and continual adaptation. 
Recent studies have demonstrated its effectiveness in improving robustness for tasks such as vision-language modeling~\cite{tan2025uncertainty} and multimodal object tracking~\cite{shao2025pura}.

Methods such as TENT~\cite{wang2020tent,mancini2018kitting,nado2020evaluating,schneider2020improving} adapt the model to distributional shifts by adjusting batch normalization statistics and updating model parameters in real time based on an entropy minimization objective. 
FSTTA~\cite{gao2023fast} performs parameter updates and restoration through two stages—gradient decomposition update and parameter decomposition recovery—effectively mitigating model instability and catastrophic forgetting caused by over-adaptation. 
PURA~\cite{shao2025pura} further extends this idea to the RGB-T tracking framework for the first time, achieving impressive results. 
However, methods involving backpropagation suffer from severe computational efficiency degradation~\cite{wang2024backpropagation}, and the presence of noise in test samples inevitably leads to unstable parameter updates and catastrophic forgetting of historical knowledge~\cite{niu2022efficient,park2025hybrid,song2023ecotta}.

Recently, cache-based TTA methods~\cite{huang2025cosmic,han2024dota} have emerged, achieving test-time optimization using only forward propagation, thereby greatly alleviating these issues. 
Tip-Adapter~\cite{zhang2022tip} first introduced the use of a key-value cache model to store historical samples and adaptively adjust predictions during testing based on cached content. 
DMN~\cite{zhang2024dual} employs a dual-memory network to separately store knowledge from training data and features from test samples. 
TDA~\cite{karmanov2024efficient} introduces the concept of a negative cache for the first time, which explicitly labels more definite missing categories within uncertain samples, thereby further reducing the impact of sample noise. 
% Since cache-based methods no longer rely on backpropagation or batch statistics updates, they effectively avoid computational inefficiency and significantly broaden the applicability of test-time adaptation.
However, since most of these methods are applied to static image processing, they lack the capability for multi-object temporal modeling.

%-------------------------------------------------------------------------

%% file: sec/3.Method.tex
\begin{figure*}[t]
    \centering
    \includegraphics[width=0.96\textwidth]{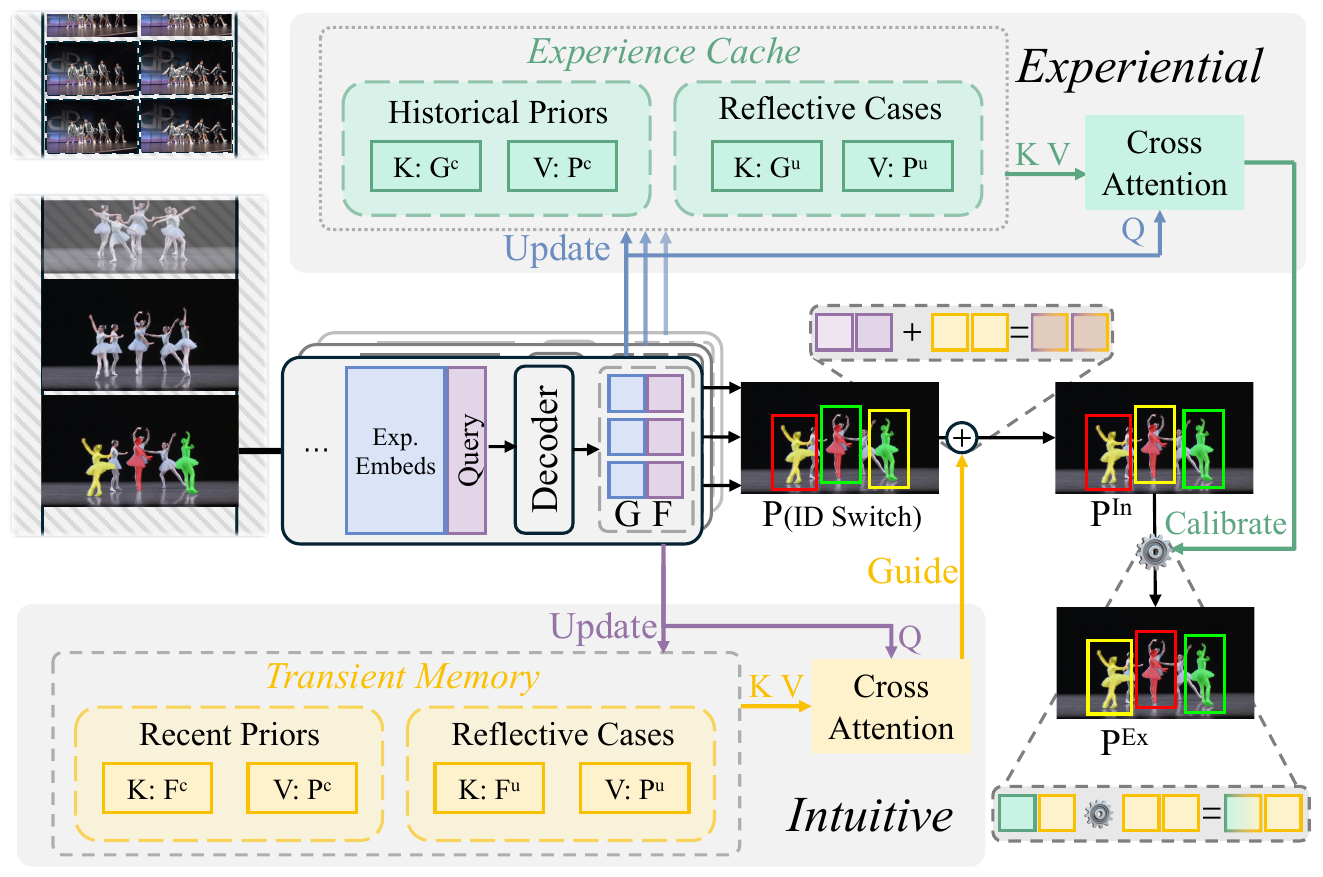}
    \caption{Overview of the proposed Test-time Calibration from Experience and Intuition (TCEI) framework. 
            The Intuitive system performs rapid inference using transient memory, while the Experiential system refines predictions with historical test experience. 
            Confident and uncertain objects are stored in caches to provide temporal priors and reflective cues. 
            “Exp. Embeds” denotes the experience embeddings, while “Query” represents the query embeddings of the Transformer decoder. 
            The experience embeddings evolve along with the query embeddings to capture object-specific characteristics.}
    \label{fig:overview}
\end{figure*}

\section{Method}
\label{sec:Method}

% \begin{figure*}
%   \centering
%   \begin{subfigure}{0.28\linewidth}
%     \fbox{\rule{0pt}{2in} \rule{.9\linewidth}{0pt}}
%     \caption{Another example of a subfigure.}
%     \label{fig:short-b}
%   \end{subfigure}
%   \caption{Example of a short caption, which should be centered.xample of a short caption, which should be centered.xample of a short caption, which should be centered.xample of a short caption, which should be centered.xample of a short caption, which should be centered.xample of a short caption, which should be centered.}
%   \label{fig:overview}
% \end{figure*}

%------------------------------o-------------------------------------------
\subsection{Overview}
\label{sec:overview}

The inference process of common MOT methods can be abstracted into a simple procedure.
Given a test image $T_{test}$, a model trained on the training data processes the image to obtain a set of detected objects $X = \{x_1, x_2, ..., x_n\}$, with corresponding object features $F = \{f_1, f_2, ..., f_n\} \in \mathbb{R}^{n \times D}$ and their predicted identities (IDs) $Y = \{y_1, y_2, ..., y_n\}$.
If the model is Transformer-based, this process typically involves feature encoding and decoding through an encoder-decoder architecture.
Here, $n$ denotes the number of objects in the current image, and $D$ represents the dimension of the object feature.
We refer to the predicted probability distribution of each object over all IDs as the prediction map $P = \{p(x_1), p(x_2), ..., p(x_n)\}$.
As shown in \cref{fig:overview},
we maintain a set of experience embeddings to capture the accumulated experience from all previously processed videos, denoted as $G = \{g_1, g_2, ..., g_n\} \in \mathbb{R}^{m \times D}$.
Here, $m$ denotes the number of experience embeddings, and each embeddings has the same dimension $D$ as the object feature.

% As shown in \cref{fig:overview}, our Intuitive system retains the model's learned knowledge from training while efficiently guide the predictions of object IDs by directly leveraging transient memory.
% The Experiential system further integrate experience derived from prior test objects, thereby reassessing and calibrating the prediction produced by the Intuitive system.
% Through this approach, our TCEI framework greatly alleviates the performance degradation caused by data distribution shifts.
% In the following \cref{sec:intuitive_system} and \cref{sec:experiential_system}, we provide detailed description of how the Intuitive system and the Experiential system function within our framework.
                             
%------------------------------o-------------------------------------------
\subsection{Intuitive System}
\label{sec:intuitive_system}

The core idea of the Intuitive system is to enhance identity association accuracy by exploiting short-term historical information to guide the current predictions.
To this end, the system leverages objects stored in transient memory to guide the model's current predictions.
First, we clarify which recent objects are valuable to be stored in transient memory and used to guide model predictions.
A straightforward strategy is to gather the object of confident predictions, denoted as $X^c$, along with their corresponding features $F^c$ and prediction map $P^c$.
These recent confident objects serve as temporal priors to guide model predictions.
Beyond this, our method also collects the object of uncertain predictions, denoted as $X^u$, along with their corresponding features $F^u$ and prediction map $P^u$.
These recent uncertain trajectory objects act as reflective cases, enabling the model to reassess its predictions and avoid making similar unreliable predictions in the current frame.
Specifically, we determine the confidence level of each object according to the entropy $E(p(x))$ of its ID prediction.
A smaller entropy value indicates a more confident prediction, whereas a larger entropy reflects higher uncertainty.
For confident objects, we select those with lower entropy values. 
In contrast, for uncertain objects, we aim to maintain their entropy value around an intermediate level $e^u$ in order to avoid including overconfident objects with extremely low entropy and noisy objects with excessively high entropy.
To achieve this, we design a transient memory mechanism with a maximum capacities of $k_c$ and $k_u$ for recent onfident and uncertain objects, respectively.
The transient memory gradually stores qualified objects while replacing outdated ones. 
Specifically, the features ($F^c$ and $F^u$) of confident and uncertain objects ($X^c$ and $X^u$) and their corresponding prediction maps ($P^c$ and $P^u$) are stored as keys and values, respectively.
To formally describe the update process, let the transient memory already contain the existing confident and uncertain object sets $X^c$ and $X^u$. 
For the incoming objects $X$ of the current frame, we extract two candidate sets 
For the current frame, we extract two candidate sets, $\Phi \subseteq \{ X^c, X \}$ and $\Psi \subseteq \{X^u, X \}$. 
The transient memory is then updated as:
\begin{equation}
\begin{aligned}
X^c_{update} &= {\arg\min_{|\Phi|= k_c}} \sum_{x \in \Phi} Entropy(x)  \\
X^u_{update} &= {\arg\min_{|\Psi|= k_u}} \sum_{x \in \Psi} |Entropy(x)-e^u|  \\
\end{aligned}
\label{eq:intuitive_update}
\end{equation}
Here, the constraints $|\Phi| = k_c$ and $|\Psi| = k_u$ ensure that the memory sizes remain unchanged after the update.
% The detailed construction process is described in ~\cite{karmanov2024efficient}.

Next, we describe how the transient memory is utilized to guide the model in predicting object IDs.
We decompose this process into two components: guidance cues and guidance strengh.
To begin with, we construct guidance cues from the prediction maps $P^{tm} = P^c \cup P^u$ of recent objects $X^{tm} = X^c \cup X^u$ in transient memory.
These cues are composed of two types: recent priors derived from confident objects and reflective cases derived from uncertain objects.
For confident objects, the prediction map is masked into a one-hot vector, where the position corresponding to the predicted ID is set to 1.
For uncertain objects, the prediction map is masked into a multi-hot vector, where all entries with prediction values greater than the threshold $\tau$ are set to -1.
All remaining entries are set to 0.
Through this process, each object naturally obtains a one-hot or multi-hot vector $V(X^{tm}) = \{\, v(x^{tm}) \mid x^{tm} \in X^{tm} \,\}$ as its guidance cues.

Subsequently, we compute the guidance strength based on the similarity between the current object features $F$ and the recent object features $F^{tm}$ stored in the transient memory.
By performing a weighted summation of the guidance cues with respect to their guidance strengths, we obtain the guidance from transient memory to the model's predictions. 
This entire process can be directly implemented through a cross-attention mechanism, formulated as:
{\small
\begin{equation}
\begin{aligned}
P^{tm} &= \mathrm{Attention}(Q = F,\, K = F^{tm},\, {V} = {V}(X^{tm})) \\
&= \mathrm{softmax}\!\left( \frac{F(F^{tm})^{\top}}{\sqrt{D}} \right) V(X^{tm})
\end{aligned}
\label{eq:intuitive_attention}
\end{equation}
}

Finally, the prediction of the Intuitive system can be formulated as:
\begin{equation}
  P^{\text{In}} = P + P^{tm}
\label{eq:intuitive_final}
\end{equation}
where P denotes the prediction map defined in \cref{sec:overview}.
The Intuitive system preserves the model's original predictions while leveraging both recent confident and uncertain objects to guide the current prediction.
Specifically, confident object provide positive guidance for accurate prediction, whereas uncertain objects act as reflective references that help the model avoid making similary unreliable predictions.

%------------------------------o-------------------------------------------
\subsection{Experiential System}
\label{sec:experiential_system}

The core idea of the Experiential system is to compensate not only for the lack of long-range temporal information in the intuitive predictions but also for their limited ability to perceive and adapt to the distribution characteristics of the testing domain. 
To this end, it leverages the historical experience accumulated during previous testing to reassess and calibrate the intuitive outputs.
Similar to the Intuitive system, both confident and uncertain objects are valuable for experience accumulation.
The confident objects serve as experience priors, while the uncertain objects act as reflective experiences.
To this end, we construct an experience cache mechanism that gradually maintains a set of confident historical objects with the lowest prediction entropy and a set of uncertain historical objects whose entropy values are closest to $e^u$. 
Their corresponding experience embeddings and prediction maps, $(G^c, P^c)$ and $(G^u, P^u)$, are stored as keys and values, respectively.
The cache is updated following the same mechanism used in the Intuitive system, as described in \cref{eq:intuitive_update}.

Next, we describe how the experience cache are used to reassess and calibrate the intuitive predictions.
Since both the transient memory and the experience cache can be used to guide model predictions, we only need to compare the extent to which each adjusts the model's outputs. 
The guidance effect of the transient memory has been detailed in the Intuitive system, and we now describe how the experience cache guides the model's predictions.
Similarly, we construct two types of guidance cues from the prediction maps of confident and uncertain objects stored in the experience cache, referred to as the experience prior cues and reflective experience cues, respectively. 
The prediction maps of confident objects are transformed into one-hot vectors, while those of uncertain objects are converted into multi-hot vectors, following the same principle as in the Intuitive system.
All guidance vectors corresponding to the objects in the experience cache $X^{ec}$ are collectively denoted as $V(X^{ec}) = \{\, v(x^{ec}) \mid x^{ec} \in X^{ec} \,\}$.
The guidance strength is then determined based on the similarity between the experience embeddings of the current objects $G$ and those stored in the experience cache $G^{ec}$. 
A weighted summation of the guidance cues according to their guidance strengths yields the guidance from the experience cache to the model's predictions.
This entire process can likewise be implemented using a cross-attention mechanism, formulated as:
\begin{equation}
\begin{aligned}
P^{ec} &= \mathrm{Attention}(Q = G,\, K = G^{ec},\, {V} = {V}(x^{ec})) \\
&= \mathrm{softmax}\!\left( \frac{G(G^{ec})^{\top}}{\sqrt{D}} \right) {V}(x^{ec}) \\
\end{aligned}
\label{eq:experiential_attention}
\end{equation}

We then use the more stable prediction guidance $P^{ec}$, obtained from the historical experience in the Experiential system, to calibrate the prediction guidance $P^{\text{In}}$ derived from the transient memory in the Intuitive system.
For the consistent components between the intuitive prediction guidance $P^{tm}$ and the experiential prediction guidance $P^{ec}$, no further modification is applied to avoid compromising the stability of model predictions.
For the inconsistent components between $P^{tm}$ and $P^{ec}$, we perform experience-based calibration.
It is worth noting that the goal of this calibration is not to make $P^{tm}$ identical to $P^{ec}$, because doing so would undermine the essence of intuitive guidance.
We emphasize that in many cases, the intuition derived from transient memory is actually the reliable choice, since recent objects share stronger associations with the current one.
To achieve this, we construct an uncertainty-based calibration mechanism.
Specifically, the uncertainty $U$ of the current objects $X$ over all IDs are computed from its prediction map $P$ as:
\begin{equation}
U = P(1 - P),
\label{eq:experiential_diff_uncertainty}
\end{equation}
Subsequently, we extract the difference between the adjustments of experiential and intuitive guidance over all IDs, formulated as:
\begin{equation}
\begin{aligned}
P^{ca} &= P^{ec} - (1- \mathrm{sim}) \cdot P^{tm},
\end{aligned}
\label{eq:experiential_diff}
\end{equation}
where $\mathrm{sim}$ denotes the similarity between their prediction adjustments over all IDs, which is computed as follows:
\begin{equation}
\mathrm{sim} = 
\frac{\left| P^{ec} - P^{tm} \right|}
    {\max\!\left( \left| P^{ec} \right|, \left| P^{tm} \right| \right)}.
\label{eq:experiential_diff_sim}
\end{equation}

Finally, the calibrated prediction produced by the Experiential system can be expressed as:
\begin{equation}
P^{\text{Ex}} = P^{\text{In}} + U \cdot P^{ca}.
\label{eq:experiential_final}
\end{equation}
This mechanism preserves the primary components of the intuitive predictions and applies moderate corrections only when the intuitive predictions are uncertain and conflict with the experiential guidance. 
By fully exploiting the complementary advantages of transient memory and long-term experience, it effectively mitigates model performance degradation under distribution shifts through comprehensive utilization of the test data.

%% file: sec/4.Experiments.tex
\section{Experiments}
\label{sec:Experiments}

\subsection{Datasets and Metrics}

{\bf Datasets.}
To specifically assess the ID prediction and association performance of TCEI, we conduct experiments on two challenging MOT datasets characterized by nonlinear motion patterns and high appearance similarity.
DanceTrack~\cite{sun2022dancetrack} is a large-scale benchmark for multi-person tracking in dance scenes, where targets frequently occlude each other, interact closely, and exhibit nearly identical appearances.
Unlike conventional MOT datasets with smooth trajectories, DanceTrack contains complex non-linear motion and frequent ID switches, making it an ideal testbed for evaluating association robustness.
SportsMOT~\cite{cui2023sportsmot} covers a wide range of competitive sports such as basketball, football, and volleyball.
It features rapid motion changes, diverse camera viewpoints, and dense player interactions, leading to highly dynamic appearance and trajectory variations.
These characteristics make SportsMOT well suited for testing the generalization and stability of association mechanisms under high-speed, high-density conditions.
% Since the MOT17~\cite{milan2016mot16} dataset is characterized by highly linear motion and a relatively small scale, and the BFT~\cite{zheng2024nettrack} dataset focuses on bird tracking, which is not the main scope of this work, we present their results in the \textbf{Supplementary Material} due to space limitations.

\noindent {\bf Metrics.}
% We follow the standard evaluation protocol to assess our method, primarily adopting the Higher Order Tracking Accuracy (HOTA) metric~\cite{luiten2021hota}, which provides a unified evaluation of detection and association performance.
% HOTA jointly considers DetA, which measures detection accuracy, and AssA, which reflects the consistency of identity associations across frames, offering a balanced assessment of overall tracking quality.
% In addition, we also report results for the IDF1 metric~\cite{ristani2016performance}, which measures the ratio of correctly identified detections over the average number of ground-truth and predicted detections, serving as a widely adopted indicator of identity preservation.
% Since our method is primarily designed to enhance association performance, we focus our analysis mainly on HOTA and AssA, as they directly capture the accuracy and robustness of cross-frame identity linking.
We follow the standard evaluation protocol and adopt HOTA~\cite{luiten2021hota} as the primary metric, as it jointly evaluates detection accuracy (DetA) and identity association accuracy (AssA).
IDF1~\cite{ristani2016performance} is also reported for completeness. 
Since our method mainly improves identity association, we focus our analysis on HOTA and AssA.

%-------------------------------------------------------------------------
\subsection{Implementation Details}

% Our method does not involve any backpropagation, making it highly lightweight and imposing minimal hardware requirements.
All our experiments are implemented in PyTorch and conducted on a single NVIDIA RTX 3090 GPU.
All hyperparameters are tuned using only the DanceTrack dataset.
In \cref{sec:intuitive_system}, the threshold $\tau$ used to generate multi-hot guidance cues from the prediction maps of uncertain targets is set to 0.03.
In both \cref{sec:intuitive_system} and \cref{sec:experiential_system}, the ideal entropy value $e^u$ for uncertain objects is set to 0.2.
Once determined, these hyperparameters remain fixed across all other datasets during evaluation.
To validate the effectiveness of our TTA approach, we adopt the simple yet representative MOTIP method~\cite{gao2025multiple} as the baseline, using only its inference stage for evaluation.

%-------------------------------------------------------------------------
\subsection{Comparison with State-of-the-art Methods}

In this section, we compare TCEI with existing methods on the DanceTrack and SportsMOT datasets. 
For Transformer-based methods, we report only the results obtained under the standard Deformable DETR and ResNet-50 frameworks to ensure fair comparison. 
It is worth noting that our proposed method is primarily designed to enhance ID prediction, i.e., association performance. 
The current state-of-the-art methods may incorporate stronger detectors, and therefore, the detection metric (DetA and IDF1) is not the primary focus of our work.

\noindent {\bf DanceTrack.}
% The highly irregular motion patterns and extremely similar appearances among targets make the DanceTrack dataset a long-standing challenge for multi-object tracking, as reliable identity association becomes particularly difficult under such conditions.
As shown in \cref{tab:compare_dance}, our method achieves 70.6\% HOTA and 63.3\% AssA on the DanceTrack test set, surpassing all existing state-of-the-art approaches across different model families.
Compared with recent Transformer-based trackers such as MOTR and MOTIP, our method improves HOTA by more than 1 percentage point and AssA by over 2 percentage points, while maintaining comparable detection accuracy (DetA).
Furthermore, when compared with the latest SSM-based trackers (e.g., MambaTrack~\cite{xiao2024mambatrack} and SambaMOTR~\cite{segu2024samba}), our approach still demonstrates a clear advantage in both overall accuracy and association quality, highlighting the effectiveness of our calibration strategy.
The performance improvement of our method can be attributed to its effective utilization of short-range and long-range historical information, which enable the model to better adapt to test data under distribution shifts.
The unusually high DetA of C-BIoU is mainly attributed to its external detector and localization-oriented design, which enhances detection quality but does not strengthen temporal association.

\begin{table}
  \caption{
    % Performance comparison with state-of-the-art methods on DanceTrack. The best performance is marked in bold. The tables are organized in ascending order of the HOTA metric.
    Performance comparison with state-of-the-art methods on DanceTrack. The best performance is marked in bold. The experimental results are obtained from reproductions based on the official code and from the official reports. The tables are organized in ascending order of the HOTA metric.
    }
  \label{tab:compare_dance}
  \centering
  \small
  \begin{tabular}{@{}l|cccc@{}}
    \toprule
    \textbf{Methods} & \textbf{HOTA} & \textbf{DetA} & \textbf{AssA} & \textbf{IDF1} \\
    \midrule
    \textit{CNN based:}             & & & & \\
    FairMOT~\cite{zhang2021fairmot} & 39.7 & 66.7 & 23.8 & 40.8 \\
    CenterTrack~\cite{zhou2020tracking} & 41.8 & 78.1 & 22.6 & 35.7 \\
    TreDeS~\cite{wu2021track}   & 43.3 & 74.5 & 25.4 & 41.2 \\
    ByteTrack~\cite{zhang2022bytetrack} & 47.7 & 71.0 & 32.1 & 53.9 \\
    GTR~\cite{zhou2022global}           & 48.0 & 72.5 & 31.9 & 50.3 \\
    QDTrack~\cite{fischer2023qdtrack}    & 54.2 & 80.1 & 36.8 & 50.4 \\
    OC-SORT~\cite{cao2023observation}   & 55.1 & 80.3 & 38.3 & 54.6 \\
    C-BIoU~\cite{yang2023hard}          & 60.6 & \textbf{81.3} & 45.4 & 61.6 \\
    \midrule
    \textit{SSM based:}             & & & & \\
    MambaTrack~\cite{xiao2024mambatrack} & 56.8 & 80.1 & 39.8 & 57.8 \\
    SambaMOTR~\cite{segu2024samba} & 67.2 & 78.8 & 57.5 & 70.5 \\
    \midrule
    \textit{Transformer based:}     & & & & \\
    TransTrack~\cite{sun2020transtrack}   & 45.5 & 75.9 & 27.5 & 45.2 \\
    MOTR~\cite{zeng2022motr}           & 54.2 & 73.5 & 40.2 & 51.5 \\
    MeMOTR~\cite{gao2023memotr}          & 63.4 & 77.0 & 52.3 & 65.5 \\
    MOTIP~\cite{gao2025multiple}      & 69.5 & 80.4 & 60.2 & 74.6 \\
    \midrule
    \textbf{TCEI (ours)}          & \textbf{70.6} & 80.2 & \textbf{62.3} & \textbf{75.6} \\
    \bottomrule
  \end{tabular}
\end{table}

\noindent {\bf SportsMOT.}
% The SportsMOT dataset features complex sports scenes with high-speed motion, frequent occlusions, and abrupt viewpoint changes, which jointly pose great challenges for maintaining consistent identity tracking.
As shown in \cref{tab:compare_sports}, , our method achieves 73.0\% HOTA and 64.0\% AssA, establishing a new state-of-the-art performance among existing approaches.
Notably, our model surpasses association-focused methods such as OC-SORT and MeMOTR, highlighting the effectiveness of our test-time calibration from experience and intuition mechanism in handling large motion variance and partial occlusions.
The improvement mainly stems from the model's ability to dynamically balance short-range adaptation and long-range calibration, thereby maintaining reliable identity consistency even under drastic motion transitions.
For fairness, we report results based on the officially released settings without incorporating any additional training data used by some prior works.
A similar trend appears on SportsMOT, with C-BIoU's reliance on an external detector yielding high DetA despite limited temporal association capability.

\begin{table}
  \caption{Performance comparison with state-of-the-art methods on the SportsMOT test set. The results are reported using the official implementations, without any additional training data.}
  \label{tab:compare_sports}
  \centering
  \small
  \begin{tabular}{@{}l|cccc@{}}
    \toprule
    \textbf{Methods} & \textbf{HOTA} & \textbf{DetA} & \textbf{AssA} & \textbf{IDF1} \\
    \midrule
    \textit{CNN based:} &&&&\\
    FairMOT~\cite{zhang2021fairmot}   & 49.3 & 70.2 & 34.7 & 53.5 \\
    GTR~\cite{zhou2022global}           & 54.5 & 64.8 & 45.9 & 55.8 \\
    QDTrack~\cite{fischer2023qdtrack}    & 60.4 & 77.5 & 47.2 & 62.3 \\
    ByteTrack~\cite{zhang2022bytetrack} & 62.1 & 76.5 & 50.5 & 69.1 \\
    CenterTrack~\cite{zhou2020tracking} & 62.7 & 82.1 & 48.0 & 60.0 \\
    OC-SORT~\cite{cao2023observation}   & 68.1 & \textbf{84.8} & 54.8 & 68.0 \\
    \midrule
    \textit{SSM based:} &&&&\\
    SambaMOTR~\cite{segu2024samba}  & 69.8 & 82.2 & 59.4 & 71.9 \\
    MambaTrack~\cite{xiao2024mambatrack} & 72.6 & 87.6 & 60.3 & 72.8 \\
    \midrule
    \textit{Transformer based:} &&&&\\
    MeMOTR~\cite{gao2023memotr}          & 68.8 & 82.0 & 57.8 & 69.9 \\
    TransTrack~\cite{sun2020transtrack}   & 68.9 & 82.7 & 57.5 & 71.5 \\
    MOTIP~\cite{gao2025multiple}      & 72.6 & 83.5 & 63.2 & 77.1\\
    \midrule
    \textbf{TCEI (ours)}          & \textbf{73.0} & 83.5 & \textbf{64.0} & \textbf{77.5} \\
    \bottomrule
  \end{tabular}
\end{table}

\noindent {\bf Comparison with Other TTA Methods.}
As shown in \cref{tab:compare_tent}, we compare the optimization results of our proposed TCEI framework with those of the conventional Tent approach~\cite{wang2020tent}.
Due to the online nature of the MOT task, where data are processed sequentially with a batch size of 1, Tent is applied only for entropy-minimization-based backpropagation, without batch normalization updates.
This setting inherently limits the applicability of most existing test-time adaptation methods, making Tent the only feasible baseline for comparison.
The results show that TCEI consistently outperforms Tent across both datasets, improving HOTA by +1.2\% and AssA by +2.2\% on DanceTrack, and achieving a +0.2\% gain in HOTA and +0.7\% in AssA on SportsMOT, all without compromising detection accuracy (DetA).
% These gains demonstrate the advantage of our test-time calibration from experience and intuition strategy, which effectively adapts model predictions through transient memory and historical experience rather than relying on direct gradient updates.
In contrast, Tent's blind backpropagation often leads to catastrophic forgetting and unstable identity associations during inference.
Moreover, TCEI achieves a notable advantage in inference efficiency over Tent, owing to the TCEI framework that operates entirely in a feed-forward manner without requiring backpropagation.
% Overall, the results provide clear evidence of the robustness and reliability of our framework in online test-time adaptation for MOT.

\begin{table}
  \caption{Comparison between Tent and our proposed TCEI framework on the DanceTrack and SportsMOT test sets. FPS indicates inference speed.}
  \label{tab:compare_tent}
  \centering
  \small
  \setlength{\tabcolsep}{4pt}         % 列间距
  \begin{tabular}{@{}llccccc@{}}
    \toprule
    \textbf{Datasets} & \textbf{Method} & \textbf{HOTA} & \textbf{DetA} & \textbf{AssA} & \textbf{IDF1} & \textbf{FPS}\\
    \midrule
    \multirow{3}{*}{DanceTrack} & No Adap. & 69.5 & \textbf{80.4} & 60.2 & 74.6 & 14\\
                                & Tent & 69.4 & 80.3 & 60.1 & 74.2 & 7\\
                                & \textbf{TCEI} & \textbf{70.6} & 80.2 & \textbf{62.3} & \textbf{75.6} & \textbf{12}\\
    % \cmidrule(lr){1-7}
    \midrule
    \multirow{3}{*}{SportsMOT} & No Adap. & 72.6 & 83.5 & 63.2 & 77.1 & 12\\
                               & Tent & 72.8 & 83.5 & 63.5 & 77.2 & 7\\
                               & \textbf{TCEI} & \textbf{73.0} & \textbf{83.5} & \textbf{64.0} & \textbf{77.5} & \textbf{9}\\
    % \cmidrule(lr){1-6}
    % \multirow{3}{*}{BFT}       & No Adap. & 71.3 & 69.2 & 73.7 & 83.1 \\
    %                            & +Tent & 71.3 & 69.2 & 73.8 & 83.1 \\
    %                            & \textbf{+TCEI} & \textbf{???} & \textbf{???} & \textbf{???} & \textbf{???} \\

    \bottomrule
  \end{tabular}
\end{table}

%-------------------------------------------------------------------------
\subsection{Ablation Studies}

% \begin{table*}
%   \caption{Domain shift}
%   \label{tab:compare_domain}
%   \centering
%   \small
%   \setlength{\tabcolsep}{10pt}         % 列间距
%   \begin{tabular}{lllcccc}
%     \toprule
%     Source & Target & Method & HOTA & DetA & AssA & IDF1 \\
%     \midrule
%     \multirow{3}{*}{DanceTrack} & \multirow{3}{*}{SportsMOT} & No Adap. & ??? & \textbf{???} & ??? & ??? \\
%                                 &                            & +Tent & ?? & ??? & ??? & ??? \\
%                                 &                            & \textbf{+TCEI} & \textbf{???} & ??? & \textbf{???} & \textbf{???} \\
%     \cmidrule(lr){1-7}
%     \multirow{3}{*}{SportsMOT} & \multirow{3}{*}{DanceTrack} & No Adap. & ??? & ??? & ??? & ??? \\
%                                &                             & +Tent & ??? & ??? & ??? & ??? \\
%                                &                             & \textbf{+TCEI} & \textbf{???} & \textbf{???} & \textbf{???} & \textbf{???} \\
%     \bottomrule
%   \end{tabular}
% \end{table*}

In this section, we conduct comprehensive ablation studies to validate the effectiveness and contribution of each component within the proposed framework.
All experiments are carried out on the DanceTrack dataset.
% , which is characterized by frequent occlusions, irregular motion patterns, and extremely high appearance similarity among objects.
% These challenging characteristics make it an ideal benchmark for examining the association improvement brought by our method.
Through these studies, we aim to provide a detailed understanding of how each module in our framework, particularly the test-time calibration from experience and intuition mechanism, contributes to the overall tracking performance.
Given that TCEI mainly targets identity association, our ablation focuses on HOTA and AssA, whereas DetA is less relevant as it reflects detection quality rather than association.

\begin{table}
  \caption{Ablation study on the Intuitive and Experiential systems conducted on the DanceTrack. The combination of both components yields the best performance, confirming their complementary contributions to association accuracy.}
  \label{tab:ablation_intuitive_experiential}
  \centering      
  \small
  \setlength{\tabcolsep}{5pt} 
  \begin{tabular}{cc cccc}
    \toprule
    \textbf{Intuitive} & \textbf{Experiential} & \textbf{HOTA} & \textbf{DetA}& \textbf{AssA} &\textbf{IDF1}\\
    \midrule
    \xmark & \xmark          & 69.5 & 80.4 & 60.2 & 74.6\\
    \cmark & \xmark          & 70.5 & 80.2 & 62.1 & 75.5\\
    \xmark & \cmark          & 70.4 & 80.3 & 61.9 & 75.6\\
    \rowcolor{gray!15}
    \cmark & \cmark          & 70.6 & 80.2 & 62.3 & 75.6\\
    \bottomrule
  \end{tabular}
\end{table}

\noindent {\bf Component Ablation.}
We conduct ablation studies to investigate the individual and combined effects of the Intuitive and Experiential systems, as summarized in \cref{tab:ablation_intuitive_experiential}.
Starting from the baseline without adaptation, introducing the Intuitive system alone improves HOTA from 69.5 to 70.5 and AssA from 60.2 to 62.1, indicating that transient memory effectively enhance the model's immediate association capability.
Similarly, incorporating only the Experiential system yields consistent gains (HOTA 70.4, AssA 61.9), verifying that historical experience can refine identity consistency across frames.
When both systems are combined, the model achieves the best overall performance (70.6 HOTA and 62.3 AssA), demonstrating the complementary nature of transient memory and long-range experience.
These results confirm that our test-time calibration from experience and intuition framework jointly leverages short-term and long-term temporal dependencies to achieve more reliable identity association.

\begin{table}
  \caption{Ablation study on the effects of confident (CO) and uncertain (UO) historical objects on the DanceTrack. “CO” and “UO” respectively denote the use of confident objects as temporal priors and uncertain objects as reflective cases. Combining both leads to the highest performance, confirming their complementary contributions to association improvement.}
  \label{tab:ablation_confident_uncertain}
  \centering      
  \small
  \begin{tabular}{cc cccc}
    \toprule
    \textbf{CO} & \textbf{UO} & \textbf{HOTA} & \textbf{DetA} & \textbf{AssA} &\textbf{IDF1}\\
    \midrule
    \xmark & \xmark          & 69.5 & 80.4 & 60.2 & 74.6\\
    \cmark & \xmark          & 69.6 & 80.3 & 60.4 & 74.6\\
    \xmark & \cmark          & 70.2 & 80.2 & 61.5 & 75.3\\
    \rowcolor{gray!15}
    \cmark & \cmark          & 70.5 & 80.2 & 62.1 & 75.6\\
    \bottomrule
  \end{tabular}
\end{table}

To further examine the roles of confident and uncertain historical objects, we conduct ablation experiments by selectively enabling each component, as shown in \cref{tab:ablation_confident_uncertain}.
When only the confident objects (CO) are utilized as historical priors, the model achieves a modest improvement over the baseline (HOTA 69.6 vs. 69.5), indicating that the cues provided by confident targets help stabilize identity assignment.
Using only the uncertain objects (UO) for reflective calibration yields a more notable gain (HOTA 70.2, AssA 61.5), suggesting that reconsidering ambiguous cases promotes better association consistency.
When both components are combined, the model attains the best overall performance (70.5 HOTA and 62.1 AssA), demonstrating that confident and uncertain historical objects provide complementary benefits, with the former offering stable temporal guidance and the latter enhancing adaptive correction under uncertainty.
% These results validate the effectiveness of incorporating both certainty- and uncertainty-driven information within the fast decision process.

\begin{figure}
  \centering
    \caption{Analysis of the maximum capacity of the confident and uncertain objects on the DanceTrack dataset.}
   \label{fig:cache_capacity}
  \includegraphics[width=0.47\textwidth]{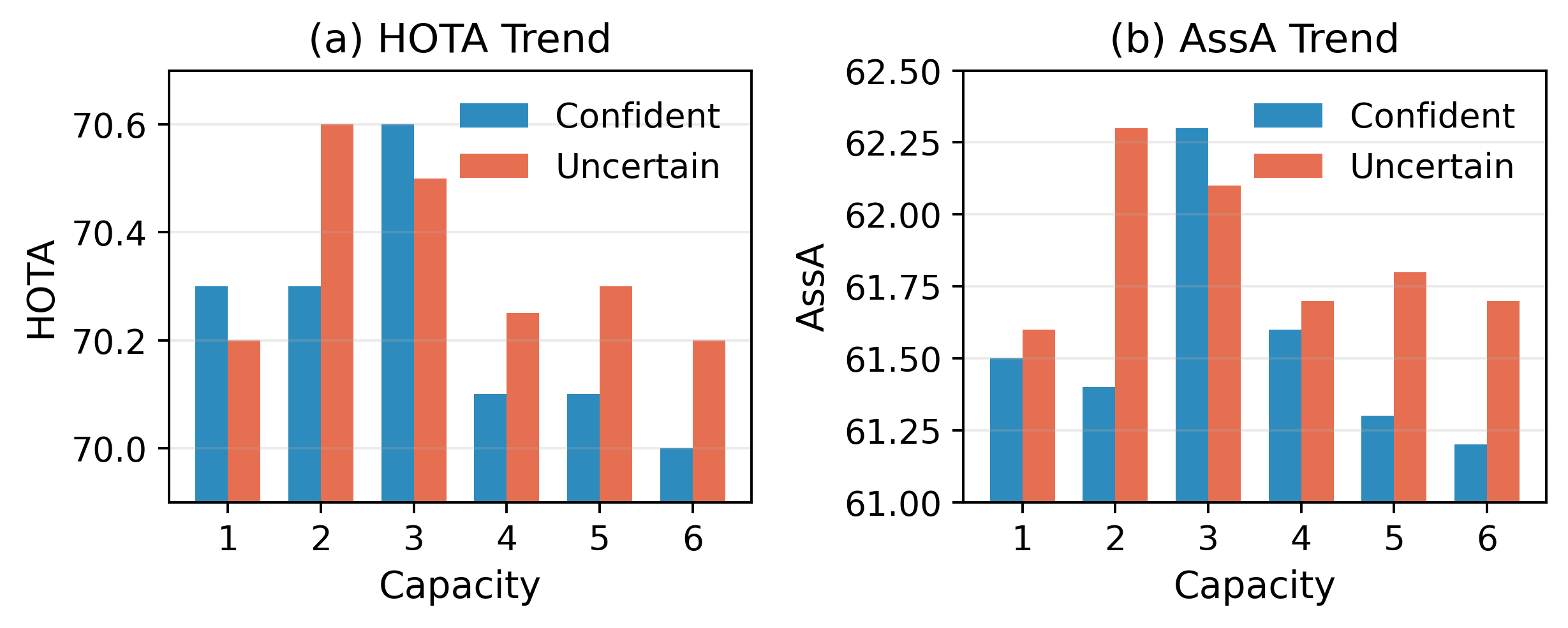}
\end{figure}

\noindent {\bf Cache Capacity Analysis.}
We further conduct a parameter study on the maximum cache capacities of the confident and uncertain objects, corresponding to $k_c$ and $k_u$ as defined in \cref{sec:intuitive_system}, to investigate the trade-off between historical object diversity and model stability.
As shown in \cref{fig:cache_capacity}, the model achieves the best performance when $k_c = 3$ and $k_u = 2$.
Performance begins to decline when the cache size deviates significantly from this configuration.
Specifically, an excessively large cache introduces highly uncertain historical objects, which may mislead subsequent optimization.
In contrast, an overly small cache limits the diversity of historical references, hindering the model's ability to achieve effective adaptation.
% These results suggest that maintaining a moderate cache size allows the model to balance stability and diversity, ensuring optimal performance during test-time adaptation.

\begin{table}
  \caption{Comparison of different calibration strategies for the Experiential system on the DanceTrack. “Average” denotes the naive averaging of Intuitive and Experiential predictions, and “Entropy” selects the prediction with lower entropy. Our selective calibration strategy, which corrects only uncertain components of the Intuitive system, achieves the best overall performance.}
  \label{tab:calibration}
  \centering
  \small
  \begin{tabular}{@{}l|cccc@{}}
    \toprule
    \textbf{Methods} & \textbf{HOTA} & \textbf{DetA} & \textbf{AssA} & \textbf{IDF1} \\
    \midrule
    No Adap. & 69.5 & \textbf{80.4} & 60.2 & 74.6\\
    Average  & 69.7 & 80.4 & 60.7 & 74.4 \\
    Entropy & 69.9 & 80.4 & 61.0 & 74.9 \\
    \midrule
    \textbf{Ours}          & \textbf{70.6} & 80.2 & \textbf{62.3} & \textbf{75.6} \\
    \bottomrule
  \end{tabular}
\end{table}

\noindent {\bf Calibration Method Analysis.}
We further investigate different calibration strategies for the Experiential system, as summarized in \cref{tab:calibration}.
In our proposed framework, the Experiential system calibrates the predictions of the Intuitive system by selectively correcting only the uncertain and erroneous components, ensuring that reliable predictions remain unaffected.
For comparison, we evaluate two alternative calibration strategies: a naive averaging approach that directly averages the adjustments of the transient memory and experience cache, and an entropy-based approach that selects the prediction with lower entropy between the two systems.
As shown in the table, both alternatives yield inferior results compared to our selective calibration strategy.
The averaging approach disrupts the predictions of both systems, while the entropy-based selection fails to handle cases where the Intuitive or Experiential system produces incorrect predictions with low entropy. 
In contrast, our method achieves the best performance, demonstrating that targeted correction of uncertain components is more effective and stable for association optimization.

%% file: sec/5.Conclusion.tex
\section{Conclusion}
To address the performance degradation of MOT models under distribution shifts between training and testing data, we propose a Test-time Calibration from Experience and Intuition framework.
The Intuitive system exploits transient memory derived from recently observed objects to guide rapid and reliable identity predictions, while the Experiential system leverages accumulated historical experience from previously processed test videos to reassess and calibrate these intuitive outputs, particularly in challenging scenarios involving appearance similarity, motion irregularity, and occlusion.
Extensive experiments across multiple MOT benchmarks demonstrate that the proposed TCEI framework for MOT substantially enhances a tracker's adaptability and robustness during online test-time inference.

%% file: sec/Acknowledgements.tex
This work was supported in part by the National Natural Science Foundation of China under Grant 62072286, 62303046, and in part by Shandong Province Graduate Education Innovation Program Project under Grant SDYKC2025223.

%% file: main.bib
@String(ICIP = {IEEE Int. Conf. Image Process.})

@String(AAAI = {AAAI})

@String(ICIP  = {ICIP})

@inproceedings{bewley2016simple,
  title={Simple online and realtime tracking},
  author={Bewley, Alex and Ge, Zongyuan and Ott, Lionel and Ramos, Fabio and Upcroft, Ben},
  booktitle={2016 IEEE international conference on image processing (ICIP)},
  pages={3464--3468},
  year={2016},
  organization={Ieee}
}

@article{welch1995introduction,
  title={An introduction to the Kalman filter},
  author={Welch, Greg and Bishop, Gary and others},
  year={1995},
  publisher={Chapel Hill, NC, USA}
}

@inproceedings{wojke2017simple,
  title={Simple online and realtime tracking with a deep association metric},
  author={Wojke, Nicolai and Bewley, Alex and Paulus, Dietrich},
  booktitle={2017 IEEE international conference on image processing (ICIP)},
  pages={3645--3649},
  year={2017},
  organization={IEEE}
}

@inproceedings{zhang2022bytetrack,
  title={Bytetrack: Multi-object tracking by associating every detection box},
  author={Zhang, Yifu and Sun, Peize and Jiang, Yi and Yu, Dongdong and Weng, Fucheng and Yuan, Zehuan and Luo, Ping and Liu, Wenyu and Wang, Xinggang},
  booktitle={European conference on computer vision},
  pages={1--21},
  year={2022},
  organization={Springer}
}

@inproceedings{cao2023observation,
  title={Observation-centric sort: Rethinking sort for robust multi-object tracking},
  author={Cao, Jinkun and Pang, Jiangmiao and Weng, Xinshuo and Khirodkar, Rawal and Kitani, Kris},
  booktitle={Proceedings of the IEEE/CVF conference on computer vision and pattern recognition},
  pages={9686--9696},
  year={2023}
}

@article{ge2021yolox,
  title={Yolox: Exceeding yolo series in 2021},
  author={Ge, Zheng and Liu, Songtao and Wang, Feng and Li, Zeming and Sun, Jian},
  journal={arXiv preprint arXiv:2107.08430},
  year={2021}
}

@inproceedings{maggiolino2023deep,
  title={Deep oc-sort: Multi-pedestrian tracking by adaptive re-identification},
  author={Maggiolino, Gerard and Ahmad, Adnan and Cao, Jinkun and Kitani, Kris},
  booktitle={2023 IEEE International conference on image processing (ICIP)},
  pages={3025--3029},
  year={2023},
  organization={IEEE}
}

@inproceedings{yang2024hybrid,
  title={Hybrid-sort: Weak cues matter for online multi-object tracking},
  author={Yang, Mingzhan and Han, Guangxin and Yan, Bin and Zhang, Wenhua and Qi, Jinqing and Lu, Huchuan and Wang, Dong},
  booktitle={Proceedings of the AAAI conference on artificial intelligence},
  volume={38},
  number={7},
  pages={6504--6512},
  year={2024}
}

@inproceedings{carion2020end,
  title={End-to-end object detection with transformers},
  author={Carion, Nicolas and Massa, Francisco and Synnaeve, Gabriel and Usunier, Nicolas and Kirillov, Alexander and Zagoruyko, Sergey},
  booktitle={European conference on computer vision},
  pages={213--229},
  year={2020},
  organization={Springer}
}

@article{zhu2020deformable,
  title={Deformable detr: Deformable transformers for end-to-end object detection},
  author={Zhu, Xizhou and Su, Weijie and Lu, Lewei and Li, Bin and Wang, Xiaogang and Dai, Jifeng},
  journal={arXiv preprint arXiv:2010.04159},
  year={2020}
}

@inproceedings{zeng2022motr,
  title={Motr: End-to-end multiple-object tracking with transformer},
  author={Zeng, Fangao and Dong, Bin and Zhang, Yuang and Wang, Tiancai and Zhang, Xiangyu and Wei, Yichen},
  booktitle={European conference on computer vision},
  pages={659--675},
  year={2022},
  organization={Springer}
}

@inproceedings{zhang2023motrv2,
  title={Motrv2: Bootstrapping end-to-end multi-object tracking by pretrained object detectors},
  author={Zhang, Yuang and Wang, Tiancai and Zhang, Xiangyu},
  booktitle={Proceedings of the IEEE/CVF conference on computer vision and pattern recognition},
  pages={22056--22065},
  year={2023}
}

@inproceedings{gao2025multiple,
  title={Multiple object tracking as id prediction},
  author={Gao, Ruopeng and Qi, Ji and Wang, Limin},
  booktitle={Proceedings of the Computer Vision and Pattern Recognition Conference},
  pages={27883--27893},
  year={2025}
}

@inproceedings{khosla2012undoing,
  title={Undoing the damage of dataset bias},
  author={Khosla, Aditya and Zhou, Tinghui and Malisiewicz, Tomasz and Efros, Alexei A and Torralba, Antonio},
  booktitle={European Conference on Computer Vision},
  pages={158--171},
  year={2012},
  organization={Springer}
}

@article{liang2025comprehensive,
  title={A comprehensive survey on test-time adaptation under distribution shifts},
  author={Liang, Jian and He, Ran and Tan, Tieniu},
  journal={International Journal of Computer Vision},
  volume={133},
  number={1},
  pages={31--64},
  year={2025},
  publisher={Springer}
}

@inproceedings{radford2021learning,
  title={Learning transferable visual models from natural language supervision},
  author={Radford, Alec and Kim, Jong Wook and Hallacy, Chris and Ramesh, Aditya and Goh, Gabriel and Agarwal, Sandhini and Sastry, Girish and Askell, Amanda and Mishkin, Pamela and Clark, Jack and others},
  booktitle={International conference on machine learning},
  pages={8748--8763},
  year={2021},
  organization={PmLR}
}

@inproceedings{mancini2018kitting,
  title={Kitting in the wild through online domain adaptation},
  author={Mancini, Massimiliano and Karaoguz, Hakan and Ricci, Elisa and Jensfelt, Patric and Caputo, Barbara},
  booktitle={2018 IEEE/RSJ International Conference on Intelligent Robots and Systems (IROS)},
  pages={1103--1109},
  year={2018},
  organization={IEEE}
}

@article{wang2020tent,
  title={Tent: Fully test-time adaptation by entropy minimization},
  author={Wang, Dequan and Shelhamer, Evan and Liu, Shaoteng and Olshausen, Bruno and Darrell, Trevor},
  journal={arXiv preprint arXiv:2006.10726},
  year={2020}
}

@article{nado2020evaluating,
  title={Evaluating prediction-time batch normalization for robustness under covariate shift},
  author={Nado, Zachary and Padhy, Shreyas and Sculley, D and D'Amour, Alexander and Lakshminarayanan, Balaji and Snoek, Jasper},
  journal={arXiv preprint arXiv:2006.10963},
  year={2020}
}

@article{schneider2020improving,
  title={Improving robustness against common corruptions by covariate shift adaptation},
  author={Schneider, Steffen and Rusak, Evgenia and Eck, Luisa and Bringmann, Oliver and Brendel, Wieland and Bethge, Matthias},
  journal={Advances in neural information processing systems},
  volume={33},
  pages={11539--11551},
  year={2020}
}

@article{gao2023fast,
  title={Fast-slow test-time adaptation for online vision-and-language navigation},
  author={Gao, Junyu and Yao, Xuan and Xu, Changsheng},
  journal={arXiv preprint arXiv:2311.13209},
  year={2023}
}

@inproceedings{shao2025pura,
  title={PURA: Parameter Update-Recovery Test-Time Adaption for RGB-T Tracking},
  author={Shao, Zekai and Hu, Yufan and Fan, Bin and Liu, Hongmin},
  booktitle={Proceedings of the Computer Vision and Pattern Recognition Conference},
  pages={22089--22098},
  year={2025}
}

@inproceedings{huang2025cosmic,
  title={COSMIC: Clique-Oriented Semantic Multi-space Integration for Robust CLIP Test-Time Adaptation},
  author={Huang, Fanding and Jiang, Jingyan and Jiang, Qinting and Li, Hebei and Khan, Faisal Nadeem and Wang, Zhi},
  booktitle={Proceedings of the Computer Vision and Pattern Recognition Conference},
  pages={9772--9781},
  year={2025}
}

@article{han2024dota,
  title={Dota: Distributional test-time adaptation of vision-language models},
  author={Han, Zongbo and Yang, Jialong and Wang, Guangyu and Li, Junfan and Xu, Qianli and Shou, Mike Zheng and Zhang, Changqing},
  journal={arXiv preprint arXiv:2409.19375},
  year={2024}
}

@inproceedings{wang2024backpropagation,
  title={Backpropagation-free network for 3d test-time adaptation},
  author={Wang, Yanshuo and Cheraghian, Ali and Hayder, Zeeshan and Hong, Jie and Ramasinghe, Sameera and Rahman, Shafin and Ahmedt-Aristizabal, David and Li, Xuesong and Petersson, Lars and Harandi, Mehrtash},
  booktitle={Proceedings of the IEEE/CVF Conference on Computer Vision and Pattern Recognition},
  pages={23231--23241},
  year={2024}
}

@inproceedings{song2023ecotta,
  title={Ecotta: Memory-efficient continual test-time adaptation via self-distilled regularization},
  author={Song, Junha and Lee, Jungsoo and Kweon, In So and Choi, Sungha},
  booktitle={Proceedings of the IEEE/CVF Conference on Computer Vision and Pattern Recognition},
  pages={11920--11929},
  year={2023}
}

@inproceedings{niu2022efficient,
  title={Efficient test-time model adaptation without forgetting},
  author={Niu, Shuaicheng and Wu, Jiaxiang and Zhang, Yifan and Chen, Yaofo and Zheng, Shijian and Zhao, Peilin and Tan, Mingkui},
  booktitle={International conference on machine learning},
  pages={16888--16905},
  year={2022},
  organization={PMLR}
}

@inproceedings{park2025hybrid,
  title={Hybrid-TTA: Continual Test-time Adaptation via Dynamic Domain Shift Detection},
  author={Park, Hyewon and Park, Hyejin and Ko, Jueun and Min, Dongbo},
  booktitle={Proceedings of the IEEE/CVF International Conference on Computer Vision},
  pages={2877--2886},
  year={2025}
}

@inproceedings{zhang2022tip,
  title={Tip-adapter: Training-free adaption of clip for few-shot classification},
  author={Zhang, Renrui and Zhang, Wei and Fang, Rongyao and Gao, Peng and Li, Kunchang and Dai, Jifeng and Qiao, Yu and Li, Hongsheng},
  booktitle={European conference on computer vision},
  pages={493--510},
  year={2022},
  organization={Springer}
}

@inproceedings{karmanov2024efficient,
  title={Efficient test-time adaptation of vision-language models},
  author={Karmanov, Adilbek and Guan, Dayan and Lu, Shijian and El Saddik, Abdulmotaleb and Xing, Eric},
  booktitle={Proceedings of the IEEE/CVF Conference on Computer Vision and Pattern Recognition},
  pages={14162--14171},
  year={2024}
}

@inproceedings{zhang2024dual,
  title={Dual memory networks: A versatile adaptation approach for vision-language models},
  author={Zhang, Yabin and Zhu, Wenjie and Tang, Hui and Ma, Zhiyuan and Zhou, Kaiyang and Zhang, Lei},
  booktitle={Proceedings of the IEEE/CVF conference on computer vision and pattern recognition},
  pages={28718--28728},
  year={2024}
}

@article{tan2025uncertainty,
  title={Uncertainty-calibrated test-time model adaptation without forgetting},
  author={Tan, Mingkui and Chen, Guohao and Wu, Jiaxiang and Zhang, Yifan and Chen, Yaofo and Zhao, Peilin and Niu, Shuaicheng},
  journal={IEEE Transactions on Pattern Analysis and Machine Intelligence},
  year={2025},
  publisher={IEEE}
}

@inproceedings{bergmann2019tracking,
  title={Tracking without bells and whistles},
  author={Bergmann, Philipp and Meinhardt, Tim and Leal-Taixe, Laura},
  booktitle={Proceedings of the IEEE/CVF international conference on computer vision},
  pages={941--951},
  year={2019}
}

@inproceedings{wojke2017simple2,
  title={Simple online and realtime tracking with a deep association metric},
  author={Wojke, Nicolai and Bewley, Alex and Paulus, Dietrich},
  booktitle={2017 IEEE international conference on image processing (ICIP)},
  pages={3645--3649},
  year={2017},
  organization={IEEE}
}

@inproceedings{hu2019joint,
  title={Joint monocular 3D vehicle detection and tracking},
  author={Hu, Hou-Ning and Cai, Qi-Zhi and Wang, Dequan and Lin, Ji and Sun, Min and Krahenbuhl, Philipp and Darrell, Trevor and Yu, Fisher},
  booktitle={Proceedings of the IEEE/CVF International Conference on Computer Vision},
  pages={5390--5399},
  year={2019}
}

@inproceedings{cui2023sportsmot,
  title={Sportsmot: A large multi-object tracking dataset in multiple sports scenes},
  author={Cui, Yutao and Zeng, Chenkai and Zhao, Xiaoyu and Yang, Yichun and Wu, Gangshan and Wang, Limin},
  booktitle={Proceedings of the IEEE/CVF international conference on computer vision},
  pages={9921--9931},
  year={2023}
}

@inproceedings{ristani2018features,
  title={Features for multi-target multi-camera tracking and re-identification},
  author={Ristani, Ergys and Tomasi, Carlo},
  booktitle={Proceedings of the IEEE conference on computer vision and pattern recognition},
  pages={6036--6046},
  year={2018}
}

@inproceedings{shin2022mm,
  title={Mm-tta: multi-modal test-time adaptation for 3d semantic segmentation},
  author={Shin, Inkyu and Tsai, Yi-Hsuan and Zhuang, Bingbing and Schulter, Samuel and Liu, Buyu and Garg, Sparsh and Kweon, In So and Yoon, Kuk-Jin},
  booktitle={Proceedings of the IEEE/CVF Conference on Computer Vision and Pattern Recognition},
  pages={16928--16937},
  year={2022}
}

@book{kahneman2011thinking,
  title={Thinking, fast and slow},
  author={Kahneman, Daniel},
  year={2011},
  publisher={macmillan}
}

@inproceedings{sun2022dancetrack,
  title={Dancetrack: Multi-object tracking in uniform appearance and diverse motion},
  author={Sun, Peize and Cao, Jinkun and Jiang, Yi and Yuan, Zehuan and Bai, Song and Kitani, Kris and Luo, Ping},
  booktitle={Proceedings of the IEEE/CVF conference on computer vision and pattern recognition},
  pages={20993--21002},
  year={2022}
}

@article{luiten2021hota,
  title={Hota: A higher order metric for evaluating multi-object tracking},
  author={Luiten, Jonathon and Osep, Aljosa and Dendorfer, Patrick and Torr, Philip and Geiger, Andreas and Leal-Taix{\'e}, Laura and Leibe, Bastian},
  journal={International journal of computer vision},
  volume={129},
  number={2},
  pages={548--578},
  year={2021},
  publisher={Springer}
}

@inproceedings{ristani2016performance,
  title={Performance measures and a data set for multi-target, multi-camera tracking},
  author={Ristani, Ergys and Solera, Francesco and Zou, Roger and Cucchiara, Rita and Tomasi, Carlo},
  booktitle={European conference on computer vision},
  pages={17--35},
  year={2016},
  organization={Springer}
}

@article{zhang2021fairmot,
  title={Fairmot: On the fairness of detection and re-identification in multiple object tracking},
  author={Zhang, Yifu and Wang, Chunyu and Wang, Xinggang and Zeng, Wenjun and Liu, Wenyu},
  journal={International journal of computer vision},
  volume={129},
  number={11},
  pages={3069--3087},
  year={2021},
  publisher={Springer}
}

@article{sun2020transtrack,
  title={Transtrack: Multiple object tracking with transformer},
  author={Sun, Peize and Cao, Jinkun and Jiang, Yi and Zhang, Rufeng and Xie, Enze and Yuan, Zehuan and Wang, Changhu and Luo, Ping},
  journal={arXiv preprint arXiv:2012.15460},
  year={2020}
}

@inproceedings{zhou2020tracking,
  title={Tracking objects as points},
  author={Zhou, Xingyi and Koltun, Vladlen and Kr{\"a}henb{\"u}hl, Philipp},
  booktitle={European conference on computer vision},
  pages={474--490},
  year={2020},
  organization={Springer}
}

@article{fischer2023qdtrack,
  title={Qdtrack: Quasi-dense similarity learning for appearance-only multiple object tracking},
  author={Fischer, Tobias and Huang, Thomas E and Pang, Jiangmiao and Qiu, Linlu and Chen, Haofeng and Darrell, Trevor and Yu, Fisher},
  journal={IEEE Transactions on Pattern Analysis and Machine Intelligence},
  volume={45},
  number={12},
  pages={15380--15393},
  year={2023},
  publisher={IEEE}
}

@inproceedings{gao2023memotr,
  title={MeMOTR: Long-term memory-augmented transformer for multi-object tracking},
  author={Gao, Ruopeng and Wang, Limin},
  booktitle={Proceedings of the IEEE/CVF International Conference on Computer Vision},
  pages={9901--9910},
  year={2023}
}

@inproceedings{yang2023hard,
  title={Hard to track objects with irregular motions and similar appearances? make it easier by buffering the matching space},
  author={Yang, Fan and Odashima, Shigeyuki and Masui, Shoichi and Jiang, Shan},
  booktitle={Proceedings of the IEEE/CVF winter conference on applications of computer vision},
  pages={4799--4808},
  year={2023}
}

@inproceedings{xiao2024mambatrack,
  title={Mambatrack: a simple baseline for multiple object tracking with state space model},
  author={Xiao, Changcheng and Cao, Qiong and Luo, Zhigang and Lan, Long},
  booktitle={Proceedings of the 32nd ACM international conference on multimedia},
  pages={4082--4091},
  year={2024}
}

@inproceedings{wu2021track,
  title={Track to detect and segment: An online multi-object tracker},
  author={Wu, Jialian and Cao, Jiale and Song, Liangchen and Wang, Yu and Yang, Ming and Yuan, Junsong},
  booktitle={Proceedings of the IEEE/CVF conference on computer vision and pattern recognition},
  pages={12352--12361},
  year={2021}
}

@inproceedings{zhou2022global,
  title={Global tracking transformers},
  author={Zhou, Xingyi and Yin, Tianwei and Koltun, Vladlen and Kr{\"a}henb{\"u}hl, Philipp},
  booktitle={Proceedings of the IEEE/CVF conference on computer vision and pattern recognition},
  pages={8771--8780},
  year={2022}
}

@article{segu2024samba,
  title={Samba: Synchronized Set-of-Sequences Modeling for Multiple Object Tracking},
  author={Segu, Mattia and Piccinelli, Luigi and Li, Siyuan and Yang, Yung-Hsu and Schiele, Bernt and Van Gool, Luc},
  journal={arXiv preprint arXiv:2410.01806},
  year={2024}
}

@inproceedings{segu2023darth,
  title={Darth: Holistic test-time adaptation for multiple object tracking},
  author={Segu, Mattia and Schiele, Bernt and Yu, Fisher},
  booktitle={Proceedings of the IEEE/CVF International Conference on Computer Vision},
  pages={9717--9727},
  year={2023}
}

@inproceedings{li2023ovtrack,
  title={Ovtrack: Open-vocabulary multiple object tracking},
  author={Li, Siyuan and Fischer, Tobias and Ke, Lei and Ding, Henghui and Danelljan, Martin and Yu, Fisher},
  booktitle={Proceedings of the IEEE/CVF conference on computer vision and pattern recognition},
  pages={5567--5577},
  year={2023}
}

@article{vaswani2017attention,
  title={Attention is all you need},
  author={Vaswani, Ashish and Shazeer, Noam and Parmar, Niki and Uszkoreit, Jakob and Jones, Llion and Gomez, Aidan N and Kaiser, {\L}ukasz and Polosukhin, Illia},
  journal={Advances in neural information processing systems},
  volume={30},
  year={2017}
}

@article{gao2017deep_relative_tracking,
  title={Deep relative tracking},
  author={Gao, Junyu and Zhang, Tianzhu and Yang, Xiaoshan and Xu, Changsheng},
  journal={IEEE Transactions on Image Processing},
  volume={26},
  number={4},
  pages={1845--1858},
  year={2017},
  publisher={IEEE}
}

@inproceedings{gao2019graph,
  title={Graph convolutional tracking},
  author={Gao, Junyu and Zhang, Tianzhu and Xu, Changsheng},
  booktitle={Proceedings of the IEEE Conference on Computer Vision and Pattern Recognition},
  pages={4649--4659},
  year={2019}
}

@article{gao2025learning,
  title={Learning Probabilistic Presence-Absence Evidence for Weakly-Supervised Audio-Visual Event Perception},
  author={Gao, Junyu and Chen, Mengyuan and Xu, Changsheng},
  journal={IEEE Transactions on Pattern Analysis and Machine Intelligence},
  volume={47},
  pages={4787 -- 4802},
  year={2025},
  publisher={IEEE}
}

@article{gao2023vectorized ,
  title={Vectorized Evidential Learning for Weakly-Supervised Temporal Action Localization},
  author={Gao, Junyu and Chen, Mengyuan and Xu, Changsheng},
  journal={IEEE transactions on pattern analysis and machine intelligence},
  volume={45},
  pages={15949 -- 15963},
  year={2023},
  publisher={IEEE}
}

@article{gao2020learning,
  title={Learning to Model Relationships for Zero-Shot Video Classification},
  author={Gao, Junyu and Zhang, Tianzhu and Xu, Changsheng},
  journal={IEEE Transactions on Pattern Analysis and Machine Intelligence},
  year={2021},
  volume={43},
  number={10},
  pages={3476--3491},
  publisher={IEEE}
}
